\documentclass[runningheads]{llncs}
\usepackage{graphicx}

\usepackage{tikz}
\usepackage{comment}
\usepackage{amsmath,amssymb} 
\usepackage{color}
\usepackage{algorithm}
\usepackage{algorithmicx}
\usepackage{algpseudocode}
\usepackage[strings]{underscore}
\usepackage{url}

\makeatletter
\def\thanks#1{\protected@xdef\@thanks{\@thanks
		\protect\footnotetext{#1}}}
\makeatother

\begin{document}
\pagestyle{headings}
\mainmatter
\def\ECCVSubNumber{6742}  

\title{Large-displacement 3D Object Tracking with Hybrid Non-local Optimization} 

\titlerunning{Large-disp. 3D Object Tracking}
%
\author{Xuhui Tian$^\dag$, Xinran Lin$^\dag$,
\thanks{$^\dag$Authors contribute equally}Fan Zhong, 
and Xueying Qin}
\authorrunning{X. Tian et al.}
%
\institute{
Shandong University, China\\
\email{zhongfan@sdu.edu.cn}
}

\maketitle

\begin{abstract}
Optimization-based 3D object tracking is known to be precise and fast, but sensitive to large inter-frame displacements. In this paper we propose a fast and effective non-local 3D tracking method. Based on the observation that erroneous local minimum are mostly due to the out-of-plane rotation, we propose a hybrid approach combining non-local and local optimizations for different parameters, resulting in efficient non-local search in the 6D pose space. In addition, a precomputed robust contour-based tracking method is proposed for the pose optimization. By using long search lines with multiple candidate correspondences, it can adapt to different frame displacements without the need of coarse-to-fine search. After the pre-computation, pose updates can be conducted very fast, enabling the non-local optimization to run in real time. Our method outperforms all previous methods for both small and large displacements. For large displacements, the accuracy is greatly improved ($81.7\% \;\text{v.s.}\; 19.4\%$). At the same time, real-time speed ($>$50fps) can be achieved with only CPU. The source code is available at \url{https://github.com/cvbubbles/nonlocal-3dtracking}.

\end{abstract}

\newcommand{\reffig}[1]{Figure~\ref{#1}}
\newcommand{\refeq}[1]{Eq.~(\ref{#1})}
\newcommand{\reftab}[1]{Table~\ref{#1}}
\newcommand{\incfig}{}

\section{Introduction}

3D object tracking aims to estimate the accurate 6-DoF pose of dynamic video objects provided with the CAD models. This is a fundamental technique for many vision applications, such as augmented reality~\cite{7368948}, robot grasping~\cite{5509171}, human-computer interaction~\cite{lepetit2005monocular}, etc. 

Previous methods can be categorized as optimization-based~\cite{1017620,prisacariu_reid_bmvc2009,stoiber2020sparse,10.1007/978-3-319-46493-0_26} and learning-based~\cite{dengPoseRBPF2021,lideepim2018,wen2021bundletrack,wen2020se}. The optimization-based methods are more efficient and more precise, while the learning-based methods are more robust by leveraging the object-specific training process and the power of GPU. Our work will focus on the optimization-based approach, aiming at mobile applications that require fast high-precision 3D tracking (e.g. augmented reality). 

\begin{figure*}[t]
	\centering
	\begin{tabular}{cc}
		\includegraphics[width=0.3\textwidth]{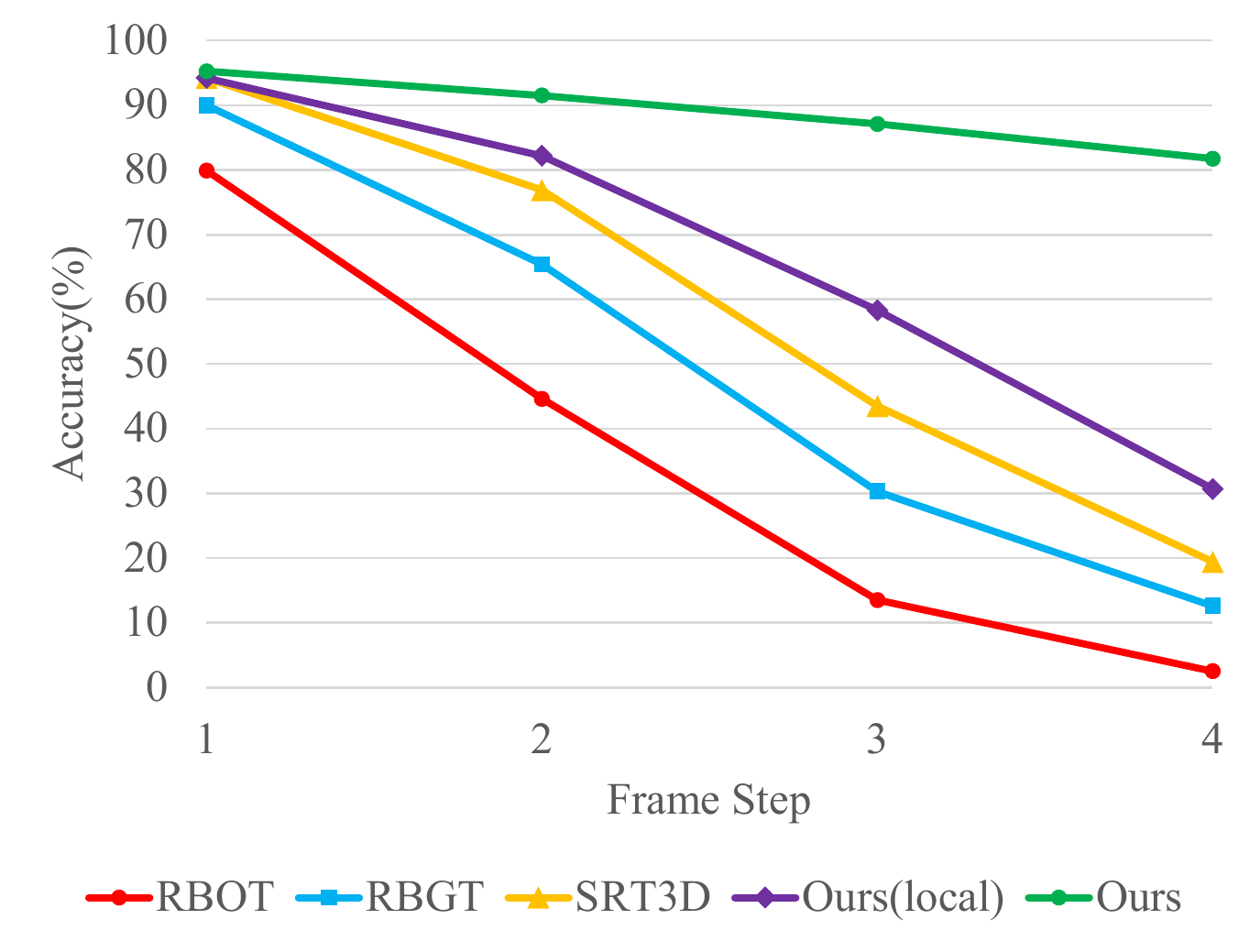} &
		\includegraphics[width=0.67\textwidth]{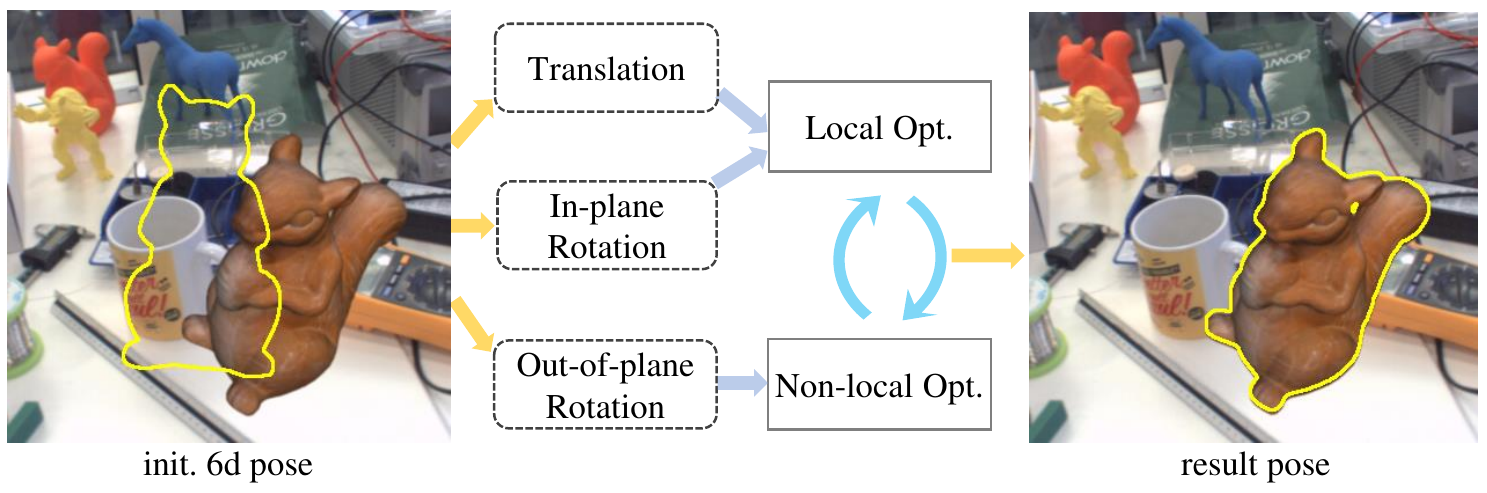} \\
		(a) & (b)
	\end{tabular}
	\caption{(a) The accuracy of previous 3D tracking methods (RBOT~\cite{tjadenRegionBasedGaussNewtonApproach2019}, RBGT~\cite{stoiber2020sparse}, SRT3D~\cite{stoiber2021srt3d})  would decrease fast with the increase of displacements (frame step $S$).  (b) The proposed hybrid non-local optimization method. 
}
	\label{fig:intro}
\end{figure*}

In order to achieve real-time speed, previous optimization-based 3D tracking methods search for only the local minima of the non-convex cost function. Note that this is based on the assumption that frame displacement is small,
which in practice is often violated due to fast object or camera movements. For large frame displacements, a good initialization is unavailable, then the local minima would deviate the true object pose. 
As shown in \reffig{fig:intro}(a), when frame displacements become large, the accuracy of previous 3D tracking methods will decrease fast.

\renewcommand{\incfig}[1]{\includegraphics[width=0.33\linewidth,height=3.3cm]{fig/1/#1.png}}

The coarse-to-fine search is commonly adopted in previous tracking methods for handling large displacements. For 3D tracking, it can be implemented by image pyramids~\cite{huangPixelWiseWeightedRegionBased2021,tjadenRegionBasedGaussNewtonApproach2019} or by varying the length of search lines~\cite{stoiber2020sparse}. However, note that since the 3D rotation is independent of the object scale in image space, coarse-to-fine search in image space would take little effect on the rotation components. 
On the other hand, although non-local tracking methods such as particle filter~\cite{kwon2013geometric} can overcome the local minimum, directly sampling in the 6D pose space would result in a large amount of computation, so previous methods~\cite{dengPoseRBPF2021,zhang2021rosefusion} always require powerful GPU to achieve real-time speed.

In this paper, we propose the first non-local 3D tracking method that can run in real-time with only CPU. \textit{Firstly}, by analyzing previous methods, we find that most tracking failures (e.g. near 90\% for SRT3D~\cite{stoiber2021srt3d})) are caused by the out-of-plane rotations. Based on this observation, we propose a hybrid approach for optimizing the 6D pose. As illustrated in \reffig{fig:intro}(b), non-local search is applied for only out-of-plane rotation, which requires to do sampling only in a 2D space instead of the original 6D pose space. An efficient search method is introduced to reduce the invocations of local joint optimizations, by pre-termination and near-to-far search. \textit{Secondly}, for better adaption to the non-local search, we propose a fast local pose optimization method that is more adaptive to frame displacements. Instead of using short search lines as in previous methods, we propose to use long search lines taking multiple candidate contour correspondences. The long search lines can be precomputed, and need not be recomputed when the pose is updated, which enables hundreds of pose update iterations to be conducted in real-time. A robust estimation method is introduced to deal with erroneous contour correspondences.  As shown in \reffig{fig:intro}(a), for the case of large displacements, our local method significantly outperforms previous methods, and the non-local method further improves the accuracy.   





\section{Related Work}

Due to space limitation, here we only briefly introduce the methods that are closely related to our work. A more comprehensive review can be found in \cite{stoiber2021srt3d}.


For textureless 3D object tracking, the optimization-based methods can be categorized as contour-based or region-based. The contour-based approaches have been studied for a long time~\cite{1017620,Harris1990RAPID,Vacchetti2004}. However, early contour-based methods are known to be sensitive to background clutters that may cause wrong contour correspondences. To solve this problem, local color information is leveraged for improving the correspondences~\cite{huangOcclusionAwareEdge2020,Seo2014}, which effectively improves accuracy.


Recent progress is mainly achieved by the region-based methods~\cite{huangPixelWiseWeightedRegionBased2021,stoiber2020sparse,stoiber2021srt3d,zhongOcclusionAwareRegionBased3D2020}. PWP3D~\cite{prisacariu_reid_bmvc2009} is the first real-time region-based method. Many methods are then proposed to improve the object segmentation. Hexner and Hagege~\cite{10.1007/s11263-015-0873-2} proposed a localized segmentation model to handle cluttered scenes. Tjaden et al.~\cite{8237285} extended the idea by using temporally consistent local color histograms. Zhong et al.~\cite{zhongOcclusionAwareRegionBased3D2020} proposed to use polar coordinates for better handling occlusion. The recent works of Stobier et al.~\cite{stoiber2020sparse,stoiber2021srt3d} proposed a sparse probabilistic model and Gaussian approximations for the derivatives in optimization, achieving state-of-the-art accuracy on the RBOT dataset~\cite{tjadenRegionBasedGaussNewtonApproach2019} and can run at a fast speed. The above methods all do only local optimization, and thus are sensitive to large displacements.



The power of deep learning can be exploited for 3D tracking when GPU is available. The 6D pose refinement network of DeepIM~\cite{lideepim2018} provides an effective way to estimate the pose difference between an input image and a rendered template. A similar network is adopted in SE(3)-TrackNet~\cite{wen2020se} for RGBD-based 3D tracking. A model-free 3D tracking method is proposed in BundleTrack~\cite{wen2021bundletrack}, by leveraging an online optimization framework. In PoseRBPF~\cite{dengPoseRBPF2021}, 6D pose tracking is formulated in a particle filter framework, with rotations searched globally in the 3D rotation space. This approach actually bridges 3D tracking with detection-based 6D pose estimation~\cite{labbe2020cosypose,pengPVNetPixelWiseVoting2019,xiangPoseCNN2018}.

\section{Adaptive Fast Local Tracking}
\label{sec:local}




To enable the non-local optimization in real-time, we first introduce a fast local optimization method that solves the local minima for arbitrary initial pose rapidly. 

\subsection{Robust Contour-based Tracking}
\label{sec:robustcontour}


As in the previous method~\cite{tjadenRegionBasedGaussNewtonApproach2019}, the rigid transformation from the model space to the camera coordinate frame is represented as:
\begin{equation}
	\mathbf{T}=\begin{bmatrix} \mathbf{R} \,& \mathbf{t} \\ \mathbf{0} \,& 1 \end{bmatrix}=\exp(\xi) \; \in \mathbb{SE}(3)
\end{equation}
with $\xi \in \mathbb{R}^6$ the parameterization of $\mathbf{T}$.

\begin{figure*}[t]
	\centering
	\begin{tabular}{ccc}

		\includegraphics[width=0.31\textwidth,height=3.3cm]{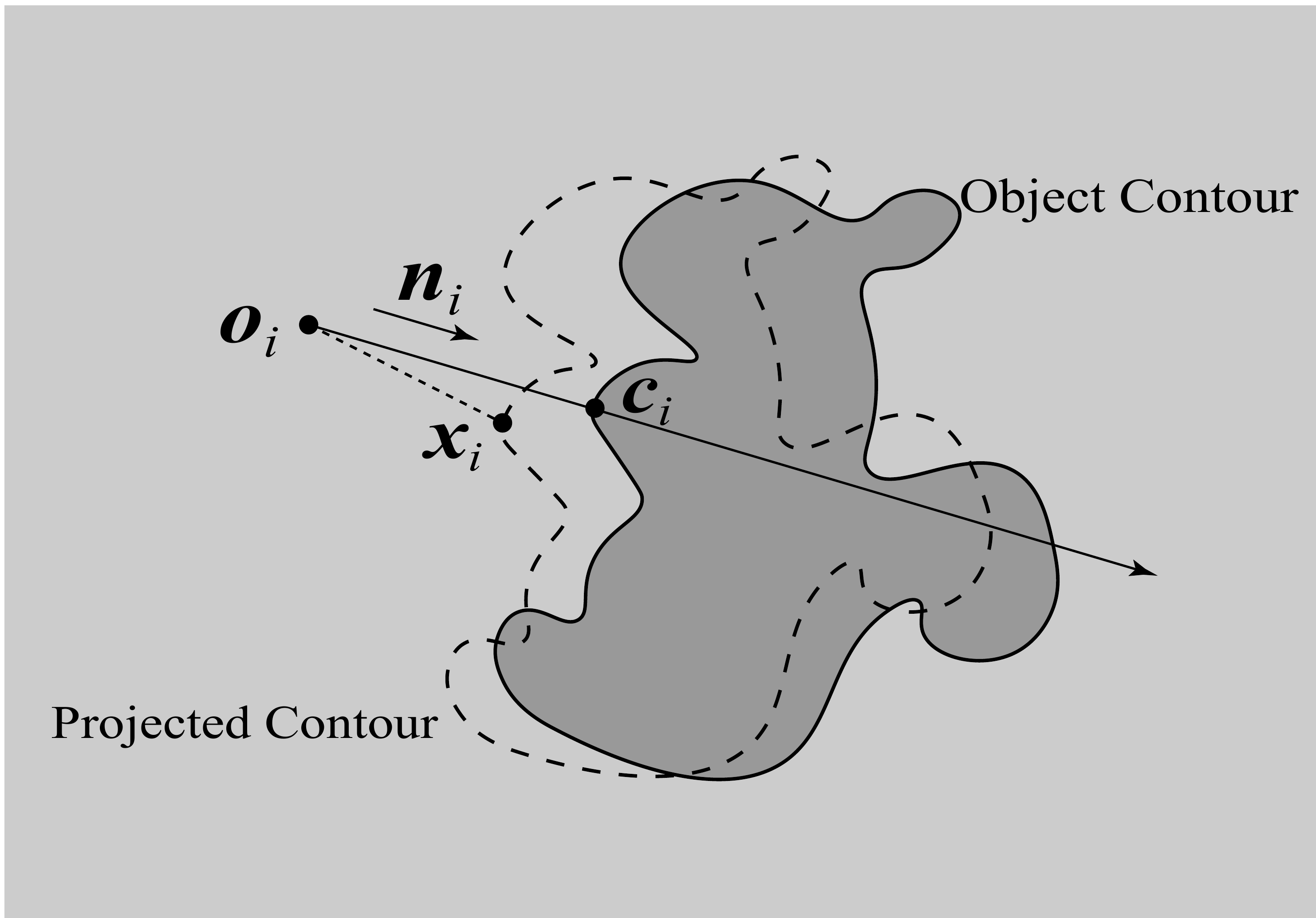} & 
		\includegraphics[width=0.31\textwidth,height=3.28cm]{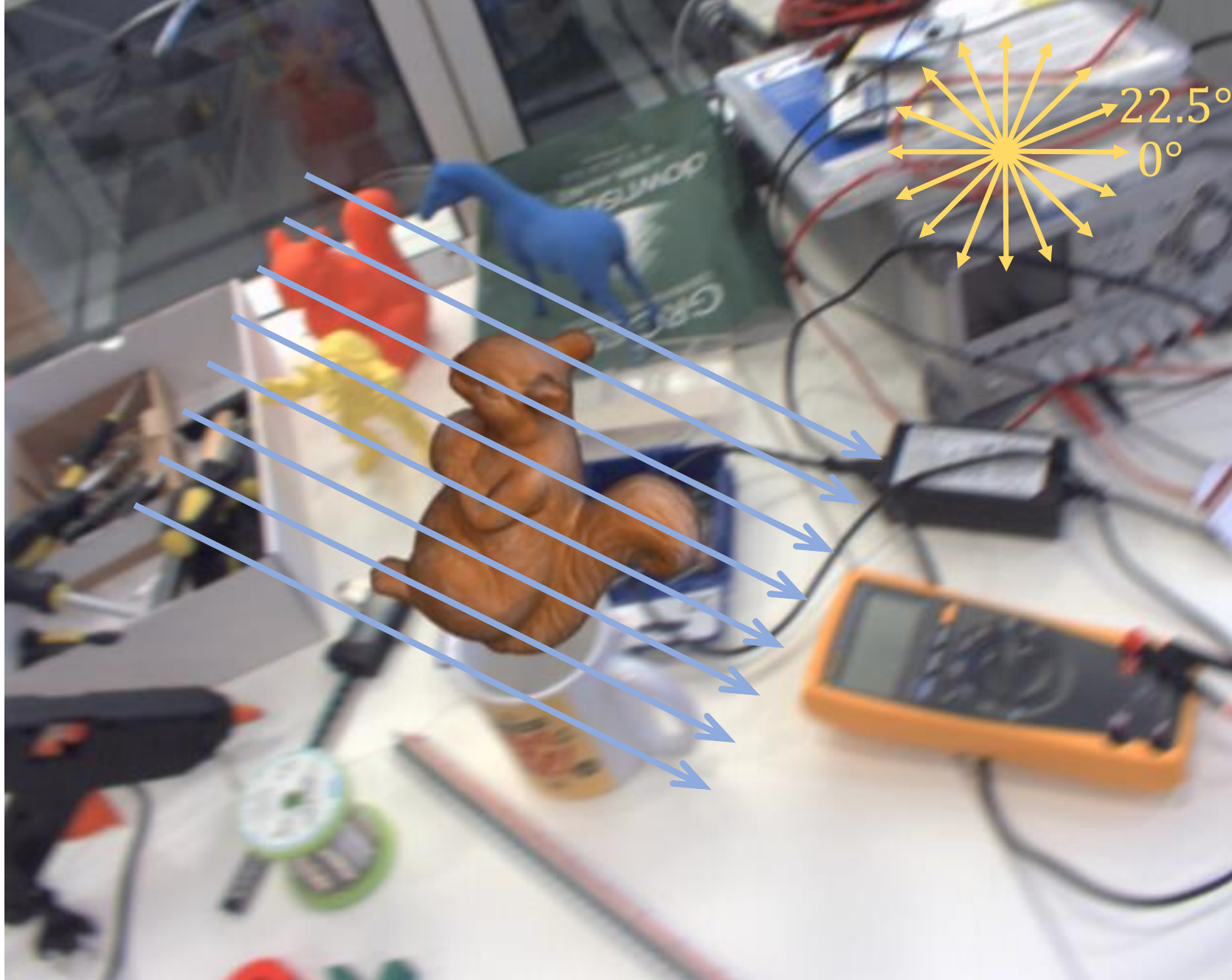}  &
		\includegraphics[width=0.31\textwidth,height=3.28cm]{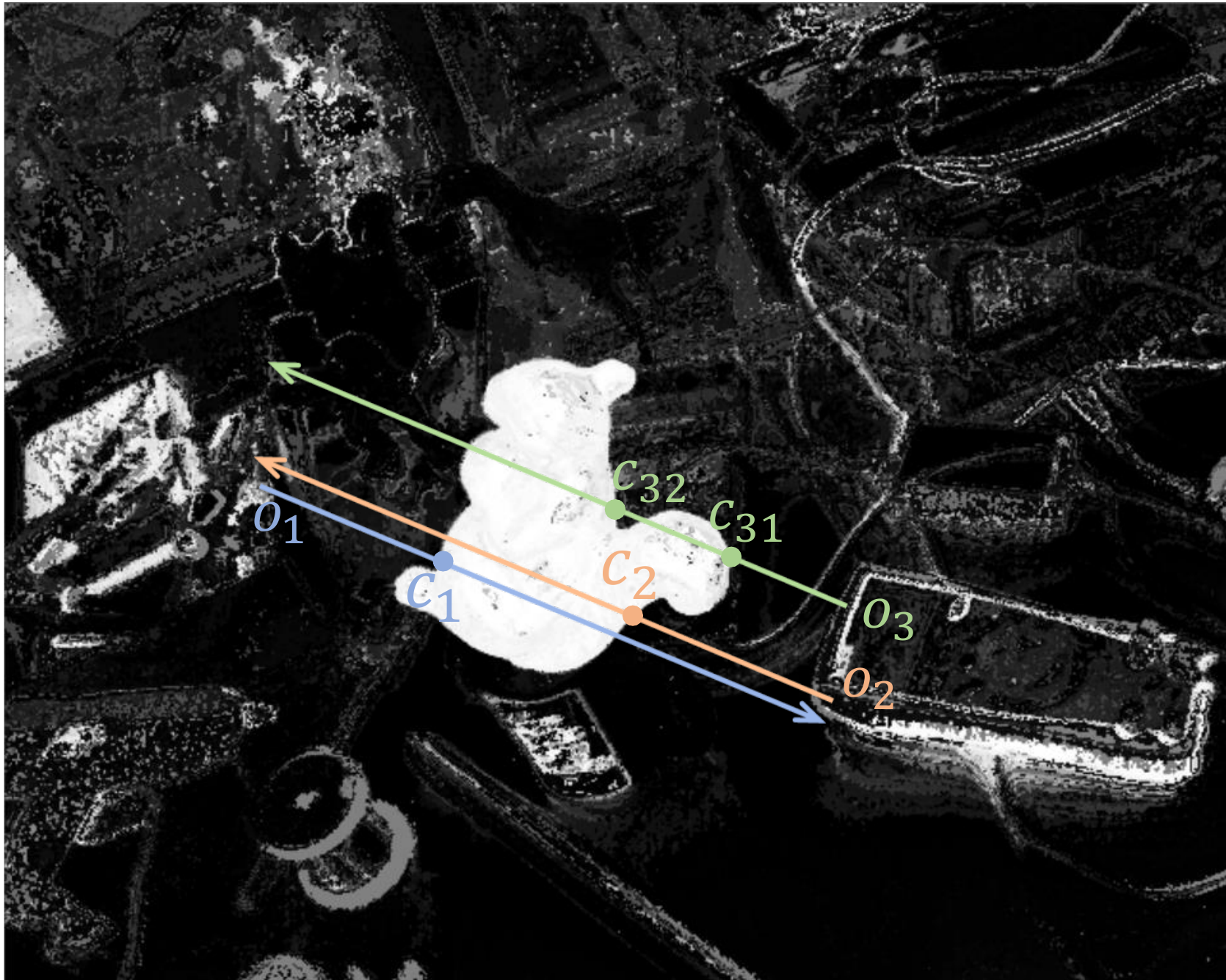} \\
		(a)  & (b) & (c)
	\end{tabular}
	\caption{(a) The adopted contour-based tracking model. (b) The search lines in one direction. The top right shows all of the directions for $D=16$. (c) The foreground probability map and exemplar search lines. Note that the contour points of $\mathbf{o}_1$ and $\mathbf{o}_2$ are on the opposite sides of the object. $\mathbf{o}_3$ has two different contour points. }
	\label{fig:scane-lines}
\end{figure*}

Given a 3D model and an initial object pose, a set of 3D model points $\mathbf{X}_i$ $(i=1,\cdots,N)$ can be sampled on the projected object contour. Denoted by $\mathbf{x}^\xi_i$ the projection of $\mathbf{X}_i$ on image plane with respect to object pose $\xi$. As illustrated in \reffig{fig:scane-lines}(a), for each $\mathbf{x}^\xi_i$, a search line $l_i=\{\mathbf{o}_i,\mathbf{n}_i\}$ is assigned, with $\mathbf{o}_i$ the start point and $\mathbf{n}_i$ the normalized direction vector. $l_i$ will be used to determine the image contour correspondence $\mathbf{c}_i$. The optimal object pose then is solved by minimizing the distance between the projected contour and the image contour:
\begin{equation}
	E(\xi)=\sum_{i=1}^{N}\omega_i\parallel\mathbf{n}_i^\top(\mathbf{x}_i^\xi-\mathbf{o}_i)-d_i\parallel^{\alpha}
	\label{eq:cost-function}
\end{equation}
with $d_i$ the distance from $\mathbf{c}_i$ to $\mathbf{o}_i$, i.e. $\mathbf{c}_i=\mathbf{o}_i+d_i\mathbf{n}_i$, so the cost function above actually measures the projected distance of $\mathbf{x}^\xi_i$ and $\mathbf{c}_i$ on the search line. $\omega_i$ is a weighting function for the $i$-th point. $\alpha$ is a constant parameter for robust estimation. $\mathbf{x}^\xi_i$ can be computed based on the pinhole camera projection function:
\begin{equation}
	\mathbf{x}_i^\xi=\pi(\mathbf{K}(\mathbf{T}\widetilde{\mathbf{X}}_i )_{3 \times 1})
	, \quad {\rm{where}}\ 
	\mathbf{K} = \begin{bmatrix}
		f_x\ & 0\ &c_x\  \\
		0\ & f_y\ &c_y\  \\
		0\ & 0\ &1\  \\
	\end{bmatrix}
\label{eq:proj}
\end{equation} 
where $\widetilde{\mathbf{X}}_i$  is the homogeneous representation of $\mathbf{X}_i$, $\pi(\mathbf{X})=[X/Z,Y/Z]^\top$ for 3D point $\mathbf{X}=[X,Y,Z]^\top$, $\mathbf{K}$ is the known $3\times3$ camera intrinsic matrix.

The above method extends previous contour-based methods~\cite{huangOcclusionAwareEdge2020,Seo2014} in the following aspects:

First, in previous methods, the search lines are generally centered at the projected point $\mathbf{x}_i^\xi$, and the image correspondences are searched in a fixed range on the two sides. This approach raises difficulty in determining the search range for the case of large displacements. In addition, since the search lines are dependent on the current object pose $\xi$, they should be recomputed once the pose is updated. The proposed search line method as shown in \reffig{fig:scane-lines}(a) can address the above problems by detaching the line configurations from $\xi$, which enables us to use long search lines 
that can be precomputed for all possible $\xi$ in the range (see Section \ref{sec:searchlines}).

Second, our method takes robust estimation with $\alpha<2$ to handle erroneous correspondences. In previous methods $\alpha=2$, so the optimization process is sensitive to the correspondence errors, and complex filtering and weighting techniques thus are necessary~\cite{huangOcclusionAwareEdge2020,Seo2014}. We will show that, by setting $\alpha$ as a small value, erroneous correspondences can be well suppressed with a simple weighting function $\omega_i$ (see Section \ref{sec:correspondence}).

\subsection{Precomputed Search Lines}
\label{sec:searchlines}

Using long search lines would make it more difficult to determine the contour correspondences. On the one hand, there may be multiple contour points on the same search line, and the number is unknown. On the other hand, due to the background clutters, searching in a larger range would be more likely to result in an error. To overcome these difficulties, for each search line $l_k$ we select multiple candidate contour points $\mathbf{c}_{km}$. Then during pose optimization, for $\mathbf{x}^\xi_i$ associated with $l_k$, $\mathbf{c}_{km}$ closest to $\mathbf{x}^\xi_i$ will be selected as the correspondence point $\mathbf{c}_i$. In this way, significant errors in $\mathbf{c}_{km}$ can be tolerated because $\mathbf{c}_{km}$ far from the projected contour would not be involved in the optimization.

For $\mathbf{x}^\xi_i$ with projected contour normal $\mathbf{\hat{n}}_i$, the associated search line can be determined as the one passing $\mathbf{x}^\xi_i$ and has direction vector $\mathbf{n}_i=\mathbf{\hat{n}}_i$. Therefore, a search line can be shared by all $\mathbf{x}^\xi_i$ on the line with the same projected normal, which enables the search lines to be precomputed. \reffig{fig:scane-lines}(b) shows all search lines in one direction. Given the ROI region containing the object, each search line will be a ray going through the ROI region. The search lines in each direction are densely arranged, so every pixel in the ROI will be associated with exactly one search line in each direction. 
%


In order to precompute all search lines, the range $[0^{\circ}, 360^{\circ})$ are uniformly divided into $D$ different directions ($D=16$ in our experiments). The search lines in each of the $D$ directions then can be precomputed. Note that a direction $o$ and its opposite direction $o+180^{\circ}$ are taken as two different directions because the contour correspondences assigned to them are different (see Section \ref{sec:correspondence}).
For arbitrary $\mathbf{x}^\xi_i$ in the ROI, there will be $D$ search lines passing through it, one of which with the direction vector closest to $\mathbf{\hat{n}_i}$ is assigned to $\mathbf{x}^\xi_i$, and then the contour correspondence $\mathbf{c}_i$ can be found as the closest candidate point. 

\subsection{Contour Correspondences based on Probability Gradients}
\label{sec:correspondence}

Correspondences would take a great effect on the resulting accuracy, so have been studied much in previous methods. Surprisingly, we find that with the proposed search line and robust estimation methods, high accuracy can be achieved with a very simple method for correspondences.

For an input image, we first compute a foreground probability map $p(\mathbf{x})$ based on color histograms. The approach is widely used in previous region-based methods~\cite{huangPixelWiseWeightedRegionBased2021,stoiber2020sparse,tjadenRegionBasedGaussNewtonApproach2019}. As shown in \reffig{fig:scane-lines}(c), the probability map is in fact a soft segmentation of the object, based on which the influence of background clutters and interior contours can be suspended effectively. Considering large object displacements, we estimate $p(\mathbf{x})$ based on the global color histograms. Note that although local color probabilities have been shown to handle complex and indistinctive color distributions better~\cite{huangPixelWiseWeightedRegionBased2021,tjadenRegionBasedGaussNewtonApproach2019}, the local window size is usually hard to be determined for the case of large displacement. 

During tracking, foreground and background color histograms are maintained in the same way as  \cite{stoiber2020sparse}, which produces probability densities $p^f(\mathbf{x})$ and $p^b(\mathbf{x})$ respectively for each pixel $\mathbf{x}$, then $p(\mathbf{x})$ is computed as: 
\begin{equation}
	p(\mathbf{x})=\frac{p^f(\mathbf{x})+\epsilon}{p^f(\mathbf{x})+p^b(\mathbf{x})+2\epsilon}
\end{equation}
where $\epsilon=10^{-6}$ is a small constant, so $p(\mathbf{x})$ would be 0.5 if $p^f(\mathbf{x})$ and $p^b(\mathbf{x})$ are both zero (usually indicating a new color has not appeared before).

Given $p(\mathbf{x})$, for each search line $l_k$, the probability value $p_k(d)$ at the pixel location $\mathbf{x}=\mathbf{o}_k+d\mathbf{n}_k$ can be resampled with bilinear interpolation. Since $p(\mathbf{x})$ is a soft segmentation of the object, the probability gradients $\nabla_k=\frac{\partial p_k(d)}{\partial d}$ can be taken as the response of object contours along $l_k$.
We thus can select the candidate contour correspondences $\mathbf{c}_{km}$ based on $\nabla_k$. Specifically, standard 1D non-maximum suppression is first applied to $\nabla_k$, then the $M$ locations with the maximum gradient response are selected as $\mathbf{c}_{km}$. 
Finally, a soft weight is computed for $\mathbf{c}_{km}$ as:
\begin{equation}
	\omega_{km}=(\frac{1}{W}\nabla_{km})^2
\end{equation}
where $W$ is a normalizing factor computed as the maximum gradient response of all candidate correspondences,
$\nabla_{km}$ is the gradient response of $\mathbf{c}_{km}$. $\omega_{km}$ will be used as the weight $\omega_i$ in \refeq{eq:cost-function} if $\mathbf{x}^\xi_i$ is matched with $\mathbf{c}_{km}$ (i.e. $\mathbf{c}_i=\mathbf{c}_{km}$).

The above method is pretty simple and elegant compared with previous methods. Note that unless there is not enough local maximum, each search line would take a fixed number of $M$ candidates ($M=3$ in our experiments). We did not even filter candidates with small responses as in usual cases. By taking a small $\alpha$, the effect of erroneous correspondences can be well suppressed.

\subsection{Pose Optimization}
\label{sec:optimization}
The cost function in \refeq{eq:cost-function} can be rewritten as
\begin{equation}
	\begin{aligned}
	E(\xi)=\sum_{i=1}^{N}\omega_i\parallel F(\xi,i)\parallel^{\alpha}, \quad {\rm{where}} \ 
	F(\xi,i)=\mathbf{n}_i^\top(\mathbf{x}_i^\xi-\mathbf{o}_i)-d_i
	\end{aligned}
\end{equation}
which can be solved with \textit{iterative reweighted least square} (IRLS) by further rewriting as
\begin{equation}
	\begin{aligned}
		E(\xi)=\sum_{i=1}^{N}\omega_i \psi_i F(\xi,i)^{2}, \quad {\rm{with}} \ \psi_i = \frac{1}{	\parallel F(\xi,i) \parallel ^{2-\alpha}}
	\end{aligned}
    \label{eq:nls}
\end{equation}
with $\psi_i$ fixed weights computed with the current $\xi$. $\psi_i$ would penalize the correspondences with larger matching residuals, which are usually caused by erroneous correspondences. Using smaller $\alpha$ can better suppress erroneous correspondences, with some sacrifice in convergence speed.

\refeq{eq:nls} is a nonlinear weighted least square problem that can be solved similarly as in previous methods~\cite{huangPixelWiseWeightedRegionBased2021,tjadenRegionBasedGaussNewtonApproach2019}. Given the Jacobian $\mathbf{J}$ of $F(\xi,i)$, the pose update of each iteration can be computed as
\begin{equation}
	\Delta \xi = -(\sum_{i=1}^N \omega_i \psi_i \mathbf{J} \mathbf{J}^{\top})^{-1}\sum_{i=1}^N \omega_i\psi_i \mathbf{J} F(\xi,i)
	\label{eq:delta}
\end{equation}
Please refer the supplementary material for details. Note that for arbitrary $\mathbf{x}^\xi_i$, the corresponding $\mathbf{c}_i$ and $d_i$ can be easily retrieved from the precomputed search lines, 
so the pose update iterations can be executed very fast.

\section{Hybrid Non-local Optimization}
\label{sec:nonlocal}

The $E(\xi)$ in \refeq{eq:cost-function} is obviously non-convex, so the method in Section \ref{sec:optimization} can obtain only the local minima. This is a common case in previous methods. Considering the complexity of the cost function and the real-time speed requirement, it is hard to be addressed with general non-convex optimization methods~\cite{jain2017non}. Here we propose an efficient non-local optimization method to address this problem.

\subsection{The Hybrid Optimization}
\label{sec:rotation}

\renewcommand{\incfig}[1]{\includegraphics[width=0.33\linewidth]{fig/5/#1}}

\begin{figure*}[t]
	\centering
	\footnotesize
	\begin{tabular}{ccc}
		\includegraphics[width=0.29\linewidth]{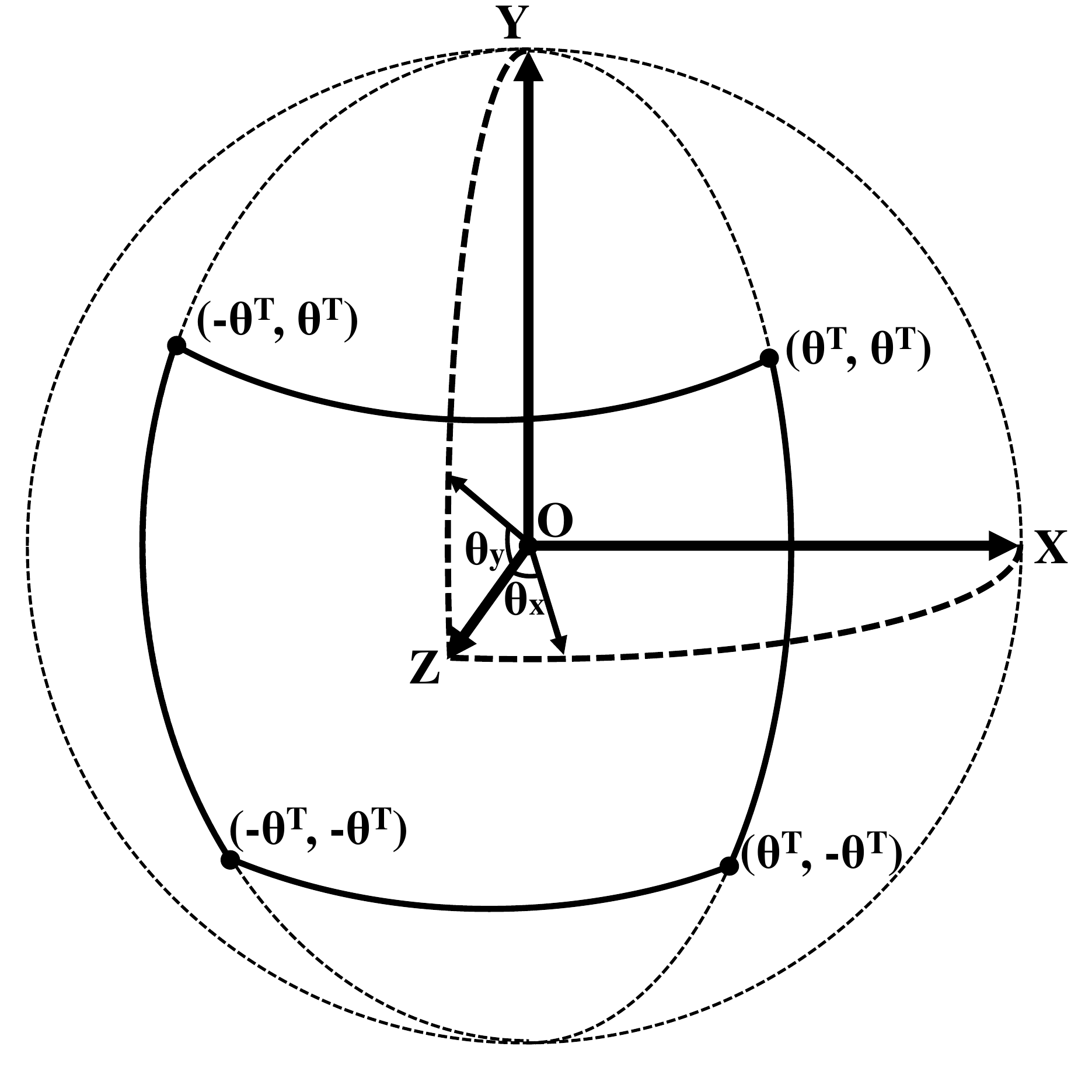} &
		\includegraphics[width=0.31\linewidth]{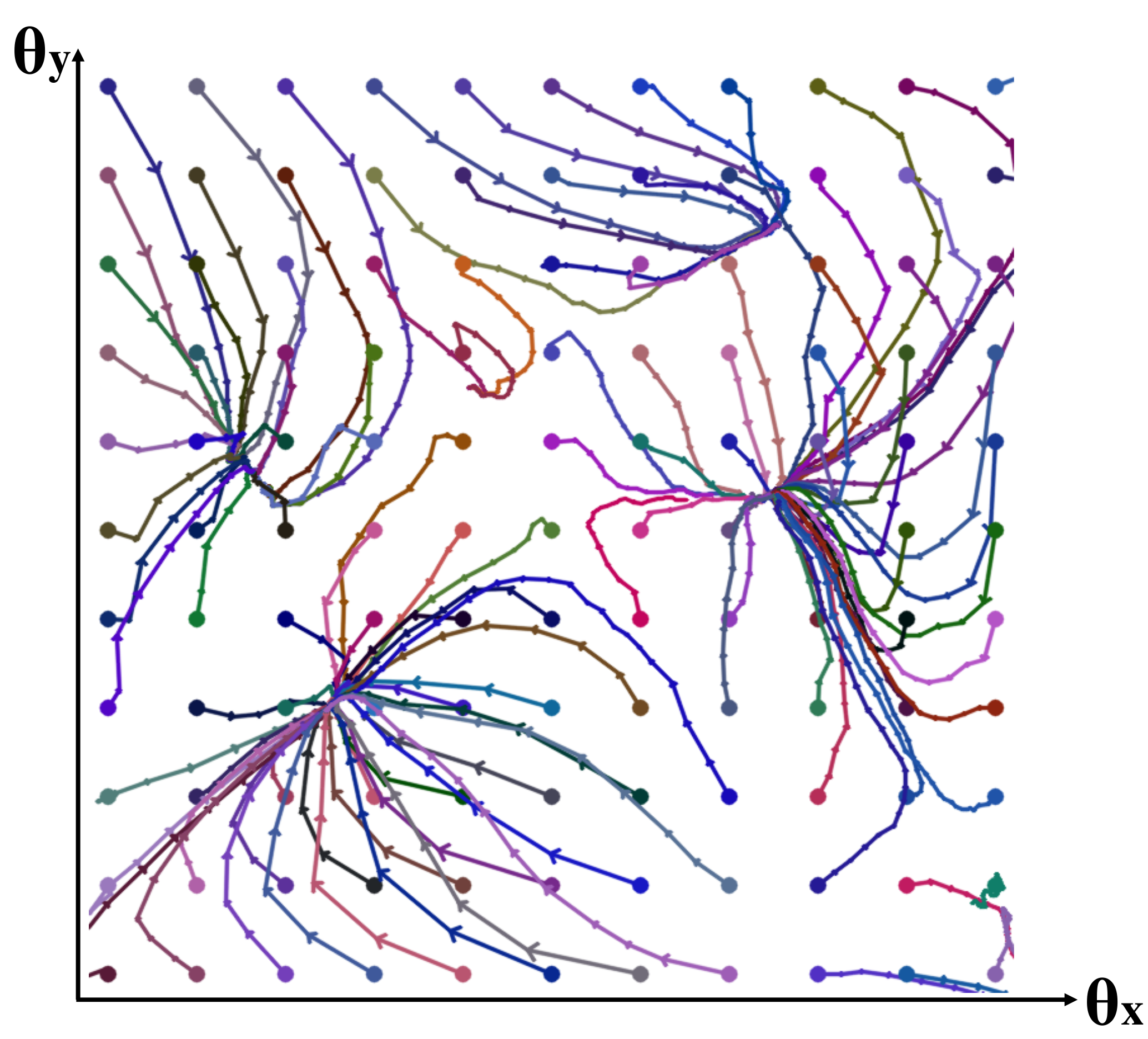} &
		\includegraphics[width=0.31\linewidth]{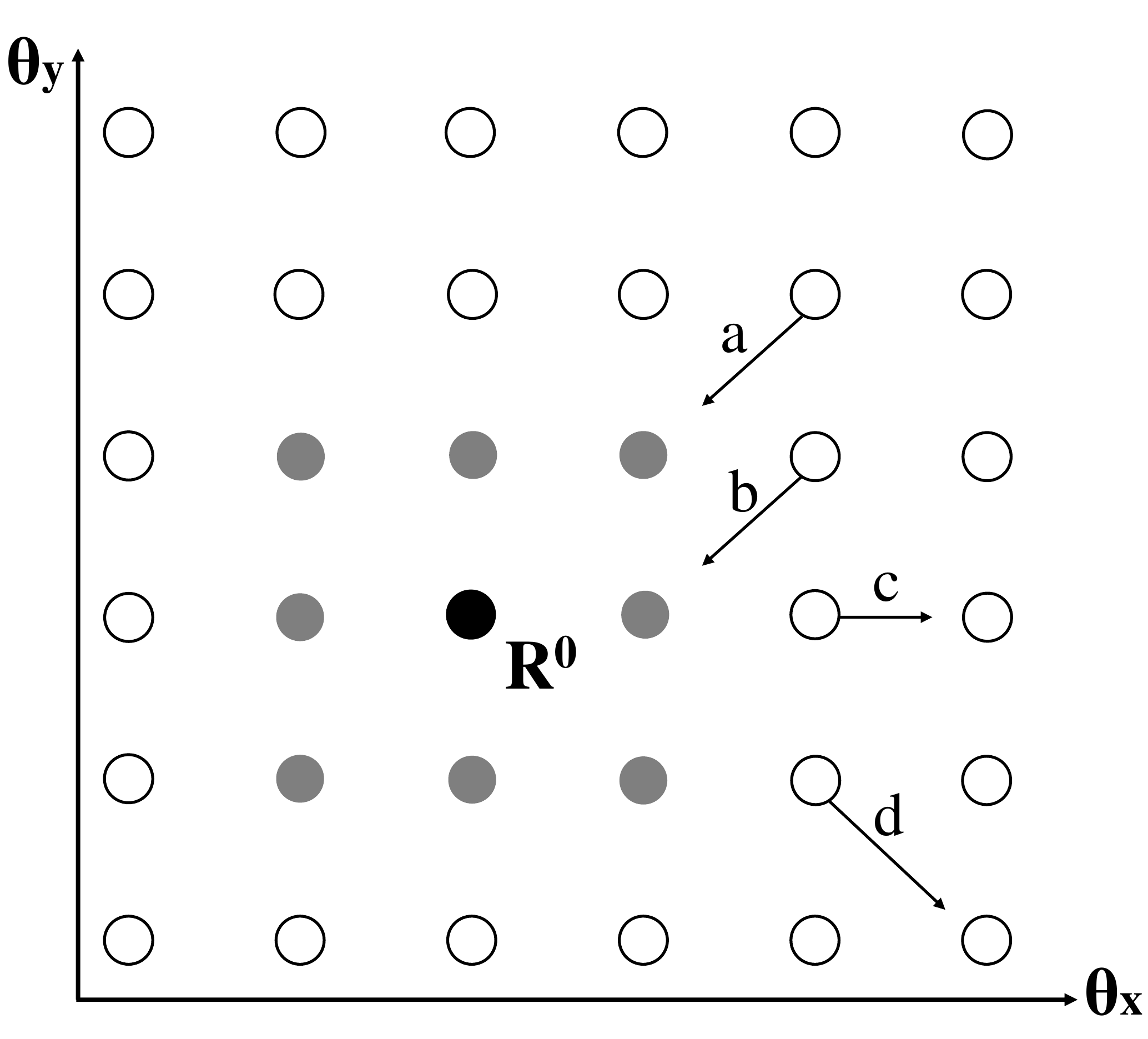} \\
		(a)  & (b) & (c)
	\end{tabular}
	\caption{(a) Parameterization of the out-of-plane rotation with $\theta_x$ and $\theta_y$. (b) Convergence paths of the naive grid search in the 2D out-of-plane parameter space, with each path corresponds to the iterations of a local optimization. (c) The near-to-far search starting from $\mathbf{R}^0$. The iterations toward $\mathbf{R}^0$ (e.g. $a,b$) may be terminated soon by the \textit{path pretermination}, while the iterations apart from $\mathbf{R}^0$ (e.g. $c,d$) may converge to other local minimum.}
	\label{fig:nonlocal}
\end{figure*}

A rotation matrix $\mathbf{R}$ can be decomposed as in-plane rotation $\mathbf{R}^{in}$ and out-of-plane rotation $\mathbf{R}^{out}$, i.e., $\mathbf{R}=\mathbf{R}^{in}\mathbf{R}^{out}$. $\mathbf{R}^{out}$ will rotate a direction vector $\mathbf{v}$ to the view axis, 
and $\mathbf{R}^{in}$ is a rotation around the view axis. By fixing an up vector, $\mathbf{R}^{out}$ can be uniquely determined from $\mathbf{v}$, then $\mathbf{R}$ can be uniquely decomposed. Please see the supplementary material for the details.

By analyzing previous tracking methods, we find that most of the erroneous local minimums are due to out-of-plane rotations. The supplementary material contains loss distributions of RBOT~\cite{tjadenRegionBasedGaussNewtonApproach2019} and SRT3D~\cite{stoiber2021srt3d}. As can be found, about 70\% of RBOT failures and 90\% of SRT3D failures are due to out-of-plane rotation. Based on this observation, we propose a hybrid approach combining non-local search and continuous local optimization, with the non-local search performed only in the 2D out-of-plane rotation space, and then jointly optimized with other parameters using local optimization.

%
%

To facilitate the non-local sampling of $\mathbf{R}^{out}$, we parameterize it as a 2D vector. Since $\mathbf{R}^{out}$ can be determined with $\mathbf{v}$, it can be parameterized in the same way. As illustrated in \reffig{fig:nonlocal}(a), each $\mathbf{v}$ around the $Z$ axis is parameterized as a 2D vector $\mathbf{\theta}=[\theta_x,\theta_y]^\top$, with $\theta_x$, $\theta_y$ $\in (-\pi/2, \pi/2)$ the elevation angles of the projections of $\mathbf{v}$ on the $XOZ$ and $YOZ$ planes, respectively. Compared with the parameterization with elevation and azimuth angles, this method can facilitate determining the neighbors in each direction and is more uniform for sampling around the $Z$ axis.


\subsection{Efficient Non-local Search}

\newcommand{\dRij}{\Delta^{R}_{ij}}

A naive non-local search method now can be easily devised based on grid search. Given an initial pose ($\mathbf{R}^0$,$\mathbf{t}^0$) and a maximum search range $\theta^T\in[0,\pi/2)$ for the out-of-plane rotation, a grid of rotation offsets $\dRij$ can be generated from the direction vectors uniformly sampled in the range $\Omega=[-\theta^T,\theta^T]\times[-\theta^T,\theta^T]$. In our implementation a fixed interval $\pi/12$ is used for the sampling. For each grid point $(i,j)$, the initial pose is reset as ($\dRij\mathbf{R}^0$,$\mathbf{t}^0$), then the local optimization method proposed in Section \ref{sec:local} is invoked to find the corresponding local minimum $\xi_{ij}$. The final pose is selected as $\xi_{ij}$ that results in smallest contour matching error $E'(\xi)=\frac{1}{N'}E(\xi)$, where $N'$ is the actual number of contour points involved in the computation of $E(\xi)$, excluding those occluded or out of image scope. 

\begin{algorithm}[t]
	\caption{Non-local 3D Tracking}
	\label{alg:NonLocalOpt}
	\begin{algorithmic}[1]
		\State \textbf{Input}: initial pose $\{\mathbf{R},\mathbf{t}\}$ and parameters $e^T$, $\theta^T$ 
		\State Precompute the search lines
		\State $\{\mathbf{R},\mathbf{t}\}$ $\gets$ $localUpdates$($\{\mathbf{R},\mathbf{t}\}$) \Comment{Please see text}
		\State $err \gets E'(\{\mathbf{R},\mathbf{t}\})$
		\If{$err > e^T$}
		\State $\mathbf{R}^0=\mathbf{R}$
		\State Sampling $\dRij$ from the range $[-\theta^T,\theta^T]\times[-\theta^T,\theta^T]$
		\State Visit grid points $(i,j)$ in breadth-first search and \textbf{do}
		\State \indent $\{\hat{\mathbf{R}},\hat{\mathbf{t}}\}$ $\gets$ $innerLocalUpdates$($\{\dRij\mathbf{R}^0,\mathbf{t}\}$) 
		\Comment{Please see text}
		
		\State \indent $\hat{err} \gets E'(\{\hat{\mathbf{R}},\hat{\mathbf{t}}\})$
		\State \indent $err, \mathbf{R},\mathbf{t}$ $\gets$ $\hat{err},\hat{\mathbf{R}},\hat{\mathbf{t}}$ \textbf{if} $\hat{err} < err$
		\State \indent Break \textbf{if}  $err < e^T$ 
		\EndIf
		\State $\{\mathbf{R},\mathbf{t}\}$ $\gets$ $localUpdates$($\{\mathbf{R},\mathbf{t}\}$)
	\end{algorithmic}	
\end{algorithm}

The above naive grid search is inefficient because the local optimizations are independent of each other, which incurs a large amount of redundant computation. 
We thus propose an improved version with the following improvements:

1) \textit{grid pre-termination}. The grid search process is pre-terminated once it finds a pose $\xi_{ij}$ that is  accurate enough, i.e., the contour matching error $E'(\xi_{ij})$ is less than a threshold $e^T$. In this way, a large amount of computation can be saved if an accurate pose is found in the early stage. Since $e^T$ is generally not easy to set manually, we adaptively estimate it as the median of the result matching errors of the previous 15 frames.

2) \textit{path pre-termination}. As visualized in \reffig{fig:nonlocal}(b), the local optimizations contain a large number of overlaps in the space of $\Omega$, which indicates that the same point in $\Omega$ may be searched multiple times. To avoid this, a new grid table that is 3 times finer than the searching grid is created for recording the visited locations in $\Omega$, then the iterations of local optimization can be pre-terminated when it reaches a visited location. Note that this may harm the optimization of other parameters (in-plane rotation and translation), so we choose to disable the pretermination for iterations with a small step in $\Omega$, which usually indicates that the current pose is close to a local minimum in $\Omega$, so the optimization should continue for further optimizing in-plane rotation and translation.

3) \textit{near-to-far search}. A near-to-far search is used instead of the sequential search in $\Omega$. As illustrated in \reffig{fig:nonlocal}(c), starting from $\mathbf{R}^0$, the locations closer to $\mathbf{R}^0$ are visited first with a breadth-first search. As a result, for the case of small displacements, $\mathbf{R}^0$ is close to the true object pose, then the search process would terminate soon with the \textit{grid pre-termination}. On the contrary, for the case of large displacements, a larger range will be searched automatically for smaller errors. This is a nice property making our method more adaptive to different displacements.

Algorithm \ref{alg:NonLocalOpt} outlines the overall procedures of the proposed non-local search method. $localUpdates$ is a procedure for local optimization, it accepts an initial pose and returns the optimized one.  $\mathbf{R}^0$,$\mathbf{t}^0$ is computed as the result of a local tracking, if $\mathbf{R}^0$,$\mathbf{t}^0$ is already accurate enough, the non-local search process would completely be skipped. Note that this is a special case of the near-to-far search process mentioned above, with \textit{grid pre-termination} activated for $\mathbf{R}^0$,$\mathbf{t}^0$. $innerLocalUpdates$ is the version of $localUpdates$ with the \textit{path pretermination}. After the non-local search, $localUpdates$ is called again to refine the pose.

\section{Implementation Details}
\label{sec:impl}

Similar to~\cite{stoiber2020sparse}, the 3D model of each object is pre-rendered as 3000 template views to avoid online rendering. The view directions are sampled uniformly with the method introduced in~\cite{arvo1992fast}. For each template, $N=200$ contour points $\mathbf{X}_i$ are sampled and stored together with their 3D surface normal. Given the pose parameter $\xi$, the image projection $\mathbf{x}^\xi_i$ as well as projected contour normals $\mathbf{n}_i$ can be computed fast with the camera projection function in Eq.(\ref{eq:proj}).


The ROI region for the current frame is determined by dilating the object bounding box of the previous frame with 100 pixels on each side. For the efficient computation of the search lines in the direction $o$, the probability map $p(\mathbf{x})$ is first rotated around the center of the ROI to align the direction $o$ with image rows. Each row of the rotated map $p^o(\mathbf{x})$ then is used for one search line. The probability gradient map $\nabla^o$ is computed from $p^o(\mathbf{x})$ using a horizontal Sobel operator with $7\times7$ kernel size. Compared with pixel difference, the smoothing process of the Sobel operator is helpful for suspending small response of contours and results in better accuracy. Note that for the opposite direction $o+180^{\circ}$, the probability gradient is negative to $\nabla^o$ at the same pixel location, and thus can be computed from $\nabla^o$ with little computation, saving nearly half of the computations for all directions.

The $localUpdates$ in Algorithm \ref{alg:NonLocalOpt} will execute the local pose optimization as in \refeq{eq:delta} up to 30 iterations, with the closest template view updated every 3 iterations. The iteration process will pre-terminate if the step $\parallel\Delta \xi\parallel$ is less than $10^{-4}$. The robust estimation parameter $\alpha$ is fixed to 0.125. Note that this is a small value for better handling erroneous correspondences. For $innerLocalUpdates$, the above settings are the same except $\alpha$, which is set as 0.75 for better convergence speed.

The non-local search range $\theta^T$ is adaptively estimated from the previous frames. When the frame $t$ is successfully tracked, the displacement with the frame $t-1$ is computed the same as the rotation error in the RBOT dataset~\cite{tjadenRegionBasedGaussNewtonApproach2019}. 
$\theta^T$ for the current frame then is computed as the median of the rotation displacements of the latest 5 frames.

\section{Experiments}
\label{sec:experiments}

In experiments we evaluated our method with the RBOT dataset~\cite{tjadenRegionBasedGaussNewtonApproach2019}, which is the standard benchmark of recent optimization-based 3D tracking methods~\cite{huangPixelWiseWeightedRegionBased2021,stoiber2020sparse,stoiber2021srt3d,zhongOcclusionAwareRegionBased3D2020}. The RBOT dataset consists of 18 different objects, and for each object, 4 sequences with different variants (i.e., \textit{regular},\textit{dynamic light}, \textit{noisy}, \textit{occlusion}) are provided. 
%
The accuracy is computed the same as in previous works with the $5cm$-$5^{\circ}$ criteria~\cite{tjadenRegionBasedGaussNewtonApproach2019}. A frame is considered as successfully tracked if the translation and rotation errors are less than $5cm$ and $5^{\circ}$, respectively. Otherwise, it will be considered as a tracking failure and the pose will be reset with the ground truth. The accuracy is finally computed as the success rate of all frames.

\newcommand{\ourslocal}{\textit{Ours}$^\text{-}$}
\newcommand{\ours}{\textit{Ours}}

\renewcommand\arraystretch{1.15}

\begin{table}[t]
	\centering
		\caption{Large-displacement tracking results on the \textit{regular} variant of the RBOT dataset \cite{tjadenRegionBasedGaussNewtonApproach2019}. In $[\;]$ is the mean and maximum displacements of rotation/translation corresponding to different frame step $S$. For $S>1$, only the methods with published code are tested. \ourslocal\; is our method without the non-local optimization(lines 5-13 in Algorithm 1). Results of other variants can be found in the supplementary material.}
	\setlength\tabcolsep{1.5pt}
	\tiny
	\begin{tabular}{lccccccccccccccccccc}
		\hline
		\rotatebox{60}{\textbf{Method}} & \rotatebox{60}{Ape} & \rotatebox{60}{Soda} & \rotatebox{60}{Vise} & \rotatebox{60}{Soup} & \rotatebox{60}{Camera} & \rotatebox{60}{Can} & \rotatebox{60}{Cat} & \rotatebox{60}{Clown} & \rotatebox{60}{Cube} & \rotatebox{60}{Driller} & \rotatebox{60}{Duck} & \rotatebox{60}{Egg Box} & \rotatebox{60}{Glue} & \rotatebox{60}{Iron} & \rotatebox{60}{Candy} & \rotatebox{60}{Lamp} & \rotatebox{60}{Phone} & \rotatebox{60}{Squirrel} & \rotatebox{60}{\textbf{Avg.}} \\
		\hline
			\noalign{\smallskip}
		\multicolumn{20}{c}{\scriptsize $S=1$ \tiny [Disp. Mean=$7.1^\circ$/15.6mm, Max=$14.8^\circ$/30.1mm]} \\
			\noalign{\smallskip}
		
		\cite{zhongOcclusionAwareRegionBased3D2020} & 88.8 & 41.3 & 94.0 & 85.9 & 86.9 & 89.0 & 98.5 & 93.7 & 83.1 & 87.3 & 86.2 & 78.5 & 58.6 & 86.3 & 57.9 & 91.7 & 85.0 & 96.2 & 82.7 \\
		\cite{huangOcclusionAwareEdge2020} & 91.9 & 44.8 & 99.7 & 89.1 & 89.3 & 90.6 & 97.4 & 95.9 & 83.9 & 97.6 & 91.8 & 84.4 & 59.0 & 92.5 & 74.3 & 97.4 & 86.4 & 99.7 & 86.9 \\
		\cite{Sun2021} & 93.0 & 55.2 & 99.3 & 85.4 & 96.1 & 93.9 & 98.0 & 95.6 & 79.5 & \textbf{98.2} & 89.7 & 89.1 & 66.5 & 91.3 & 60.6 & 98.6 & 95.6 & 99.6 & 88.1 \\
		\cite{huangPixelWiseWeightedRegionBased2021} & 94.6 & 49.4 & 99.5 & 91.0 & 93.7 & 96.0 & 97.8 & 96.6 & 90.2 & \textbf{98.2} & 93.4 & 90.3 & 64.4 & 94.0 & 79.0 & \textbf{98.8} & 92.9 & 99.8 & 89.9 \\
			\noalign{\smallskip}
		\cite{tjadenRegionBasedGaussNewtonApproach2019} & 85.0 & 39.0 & 98.9 & 82.4 & 79.7 & 87.6 & 95.9 & 93.3 & 78.1 & 93.0 & 86.8 & 74.6 & 38.9 & 81.0 & 46.8 & 97.5 & 80.7 & 99.4 & 79.9 \\
		\cite{stoiber2020sparse} & 96.4 & 53.2 & 98.8 & 93.9 & 93.0 & 92.7 & 99.7 & 97.1 & 92.5 & 92.5 & 93.7 & 88.5 & 70.0 & 92.1 & 78.8 & 95.5 & 92.5 & 99.6 & 90.0 \\
		\cite{stoiber2021srt3d} & 98.8 & 65.1 & 99.6 & 96.0 & \textbf{98.0} & \textbf{96.5} & \textbf{100} & 98.4 & 94.1 & 96.9 & 98.0 & 95.3 & 79.3 & 96.0 & 90.3 & 97.4 & \textbf{96.2} & 99.8 & 94.2 \\
		\ourslocal & \textbf{99.8} & 65.6 & 99.5 & 95.0 & 96.6 & 92.6 & \textbf{100} & 98.7 & 95.0 & 97.1 & 97.4 & 96.1 & 83.3 & 96.9 & 91.5 & 95.8 & 95.2 & 99.7 & 94.2 \\
		\ours & \textbf{99.8} & \textbf{67.1} & \textbf{100} & \textbf{97.8} & 97.3 & 93.7 & \textbf{100} & \textbf{99.4} & \textbf{97.4} & 97.6 & \textbf{99.3} & \textbf{96.9} & \textbf{84.7} & \textbf{97.7} & \textbf{93.4} & 96.7 & 95.4 & \textbf{100} & \textbf{95.2} \\
		\hline
			\noalign{\smallskip} 
		\multicolumn{20}{c}{\scriptsize $S=2$ \tiny [Disp. Mean=$14.0^\circ$/30.7mm, Max=$28.1^\circ$/57.8mm]} \\
			\noalign{\smallskip} 
		\cite{tjadenRegionBasedGaussNewtonApproach2019} & 37.6 & 11.4 & 72.0 & 46.6 & 45.2 & 44.0 & 46.6 & 52.0 & 24.6 & 65.6 & 46.4 & 44.6 & 13.6 & 42.4 & 22.8 & 67.8 & 45.2 & 75.2 & 44.6 \\
		\cite{stoiber2020sparse} & 83.4 & 21.8 & 72.4 & 75.0 & 68.4 & 58.2 & 86.4 & 78.8 & 74.0 & 58.0 & 80.4 & 65.4 & 38.8 & 63.8 & 41.6 & 59.4 & 61.8 & 90.0 & 65.4 \\
		\cite{stoiber2021srt3d} & 94.0 & 30.6 & 82.8 & 83.4 & 78.0 & 72.8 & 90.2 & 90.0 & 81.8 & 72.2 & 90.6 & 77.6 & 56.4 & 79.0 & 62.6 & 70.8 & 76.4 & 94.4 & 76.9 \\
		\ourslocal & 97.2 & 38.4 & 94.6 & 85.8 & 87.2 & 78.2 & 91.4 & 92.6 & 84.6 & 82.8 & 93.6 & 82.2 & 61.6 & 87.4 & 66.8 & 77.4 & 78.8 & 98.2 & 82.2 \\
		\ours & \textbf{100} & \textbf{49.0} & \textbf{99.4} & \textbf{96.8} & \textbf{94.4} & \textbf{90.2} & \textbf{99.6} & \textbf{99.4} & \textbf{95.2} & \textbf{93.2} & \textbf{98.8} & \textbf{92.6} & \textbf{72.0} & \textbf{95.4} & \textbf{88.0} & \textbf{93.4} & \textbf{89.8} & \textbf{100} & \textbf{91.5} \\
		\hline
			\noalign{\smallskip} 
		\multicolumn{20}{c}{\scriptsize $S=3$ \tiny [Disp. Mean=$20.8^\circ$/45.8mm, Max=$39.5^\circ$/81.0mm]} \\
			\noalign{\smallskip} 
		\cite{tjadenRegionBasedGaussNewtonApproach2019} & 8.1 & 0.3 & 28.8 & 9.0 & 12.3 & 9.9 & 14.1 & 16.8 & 4.8 & 19.2 & 17.1 & 11.7 & 2.4 & 12.9 & 3.9 & 22.2 & 11.1 & 37.8 & 13.5 \\
		\cite{stoiber2020sparse} & 47.4 & 7.2 & 26.4 & 35.7 & 28.5 & 17.4 & 43.2 & 42.0 & 40.5 & 25.2 & 47.1 & 30.3 & 12.0 & 31.2 & 14.1 & 24.6 & 25.2 & 45.9 & 30.2 \\
		\cite{stoiber2021srt3d} & 70.6 & 12.6 & 42.6 & 48.9 & 41.4 & 30.6 & 54.4 & 55.9 & 53.2 & 38.7 & 63.7 & 43.2 & 27.9 & 44.1 & 22.8 & 36.6 & 35.7 & 59.2 & 43.5 \\
		\ourslocal & 81.7 & 15.9 & 69.1 & 68.2 & 55.3 & 44.7 & 65.5 & 74.8 & 68.5 & 53.8 & 81.4 & 51.7 & 34.5 & 68.8 & 35.1 & 41.7 & 50.8 & 88.3 & 58.3 \\
		\ours & \textbf{99.4} & \textbf{37.8} & \textbf{99.4} & \textbf{94.9} & \textbf{91.6} & \textbf{83.5} & \textbf{99.4} & \textbf{98.5} & \textbf{93.1} & \textbf{84.7} & \textbf{99.7} & \textbf{87.4} & \textbf{62.2} & \textbf{91.6} & \textbf{79.3} & \textbf{85.3} & \textbf{80.2} & \textbf{100} & \textbf{87.1} \\
		\hline
			\noalign{\smallskip}
		\multicolumn{20}{c}{\scriptsize $S=4$ \tiny [Disp. Mean=$27.7^\circ$/60.7mm, Max=$54.6^\circ$/97.6mm]} \\
			\noalign{\smallskip} 
		\cite{tjadenRegionBasedGaussNewtonApproach2019} & 0.8 & 0.0 & 5.6 & 2.0 & 3.2 & 2.4 & 3.6 & 1.6 & 0.4 & 5.2 & 2.4 & 2.8 & 0.4 & 3.2 & 0.4 & 2.8 & 1.2 & 7.6 & 2.5 \\
		\cite{stoiber2020sparse} & 22.0 & 2.0 & 10.0 & 14.8 & 12.4 & 4.8 & 18.4 & 17.2 & 16.4 & 10.8 & 22.0 & 14.0 & 4.8 & 12.8 & 5.2 & 8.4 & 10.4 & 19.6 & 12.6 \\
		\cite{stoiber2021srt3d} & 36.4 & 3.2 & 15.2 & 22.4 & 17.2 & 10.0 & 26.0 & 26.0 & 28.0 & 16.8 & 37.6 & 18.0 & 10.4 & 19.2 & 6.8 & 12.8 & 16.4 & 27.2 & 19.4 \\
		\ourslocal & 50.0 & 8.0 & 34.8 & 42.0 & 28.8 & 15.2 & 36.4 & 41.2 & 39.2 & 22.0 & 55.2 & 24.8 & 12.4 & 38.0 & 10.0 & 16.4 & 22.0 & 56.0 & 30.7 \\
		\ours & \textbf{96.8} & \textbf{31.6} & \textbf{97.6} & \textbf{93.6} & \textbf{85.6} & \textbf{77.6} & \textbf{98.0} & \textbf{97.6} & \textbf{84.8} & \textbf{80.0} & \textbf{98.8} & \textbf{82.0} & \textbf{46.8} & \textbf{86.0} & \textbf{60.0} & \textbf{80.8} & \textbf{74.0} & \textbf{99.6} & \textbf{81.7} \\
		\hline
	\end{tabular}

	\label{tab:LargeDisplacementTracking}
\end{table}

\subsection{Comparisons}
\label{sec:comparisons}

Table \ref{tab:LargeDisplacementTracking} compares the accuracy of our method with previous methods, including RBOT~\cite{tjadenRegionBasedGaussNewtonApproach2019}, RBGT~\cite{stoiber2020sparse}, SRT3D~\cite{stoiber2021srt3d}, etc. The cases of large displacements are tested with different frame step $S$. Each sequence of RBOT dataset contains 1001 frames, from which a sub-sequence $\{0,S,2S,3S,\cdots\}$ is extracted for given frame step $S$. The mean and maximum displacements for different $S$ are included in Table \ref{tab:LargeDisplacementTracking}. Note that the objects in RBOT actually move very fast, so $S=4$ is indeed very challenging. For the compared methods, the results of $S=1$ are from the original paperw, and the results of $S>1$ are computed with the authors' code with default parameters. Therefore, only the methods with published code are tested for $S>1$. For our method, the same setting is used for all $S$.

As is shown, our method constantly outperforms previous methods for different frame steps. More importantly, the margin of accuracy becomes larger and larger with the increase of frame step, showing the effectiveness of our method in handling large displacements. When $S=4$, the average frame displacement is $27.7^{\circ}/60.7mm$, which is hard for local tracking methods. In this case, the most competitive method SRT3D~\cite{stoiber2021srt3d} can achieve only $19.4\%$ accuracy, while our method still obtains $81.7\%$ accuracy. Note that this accuracy is even higher than the accuracy of RBOT~\cite{tjadenRegionBasedGaussNewtonApproach2019} for $S=1$ ($79.9\%$). We have also tested with stricter $2cm$-$2^{\circ}$ criteria and the tracking accuracy is $84.1\%$. As a comparison, in this case, SRT3D~\cite{stoiber2021srt3d} achieves only $78.7\%$ accuracy.


Table \ref{tab:LargeDisplacementTracking} also shows the results of the proposed local tracking method (\ourslocal). As can be found, our local method still significantly outperforms previous methods for large displacements. This is mainly attributed to the new search line model, which makes our method more adaptive to displacements. Note that our method does not even use coarse-to-fine search, while the compared methods all exploit coarse-to-fine search for handling large displacements.


\renewcommand{\incfig}[1]{\includegraphics[width=0.19\linewidth]{fig/vis/#1}}

\begin{figure*}[t]
	\centering
	\scriptsize
	\begin{tabular}{ccccc}
		\incfig{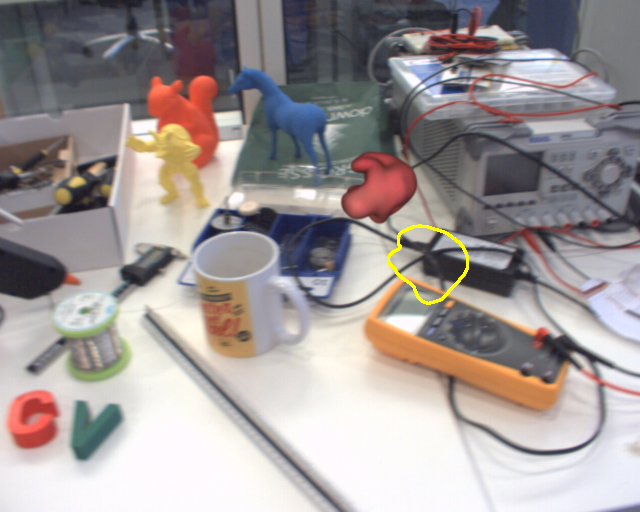} & \incfig{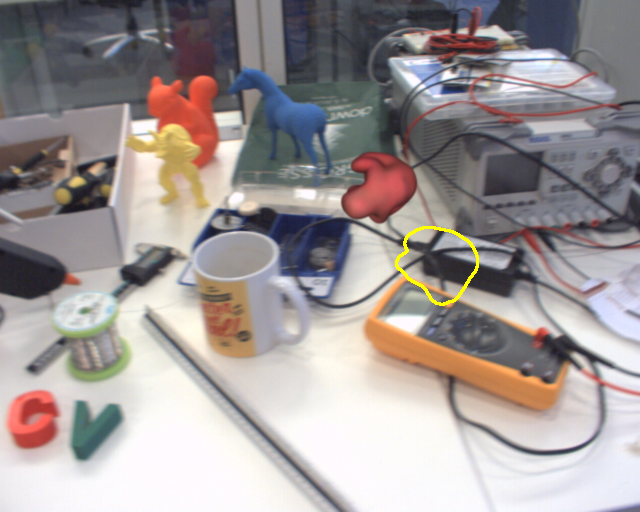} & \incfig{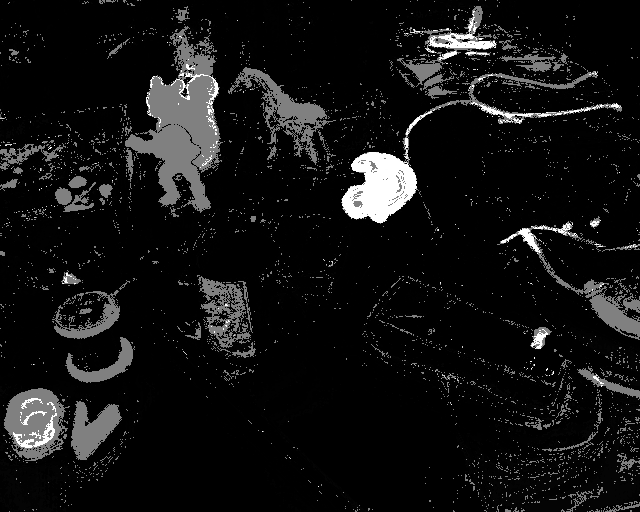} & \incfig{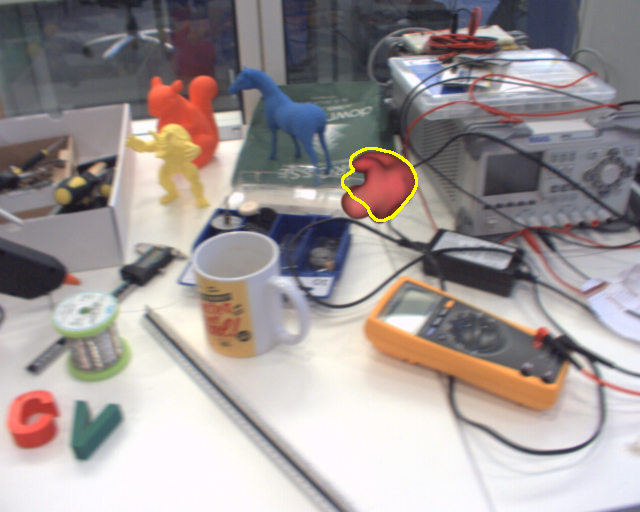} & \incfig{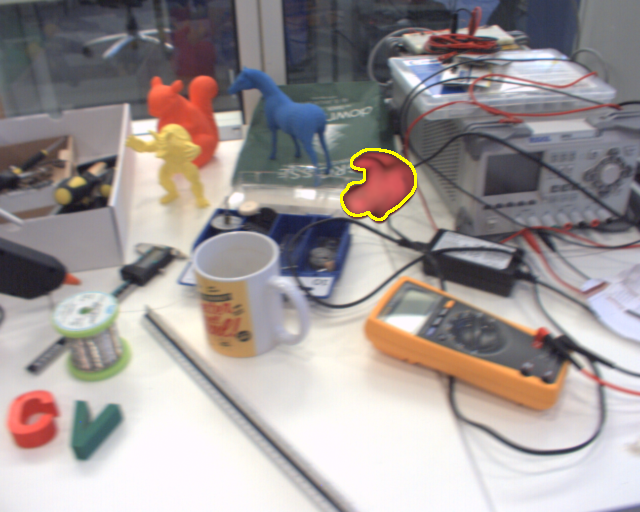}   \\
		\incfig{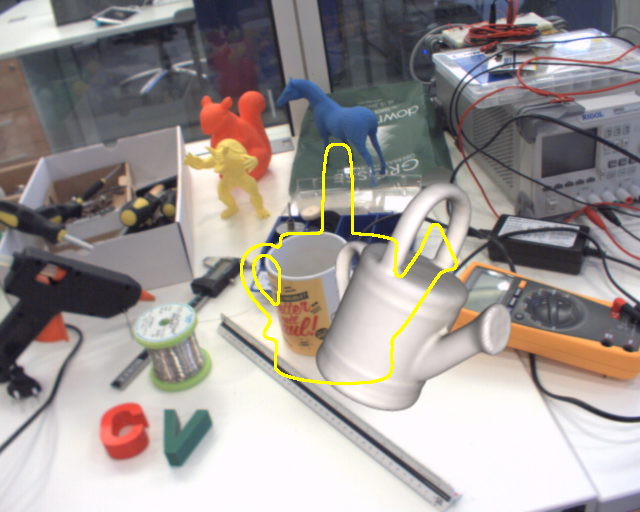} & \incfig{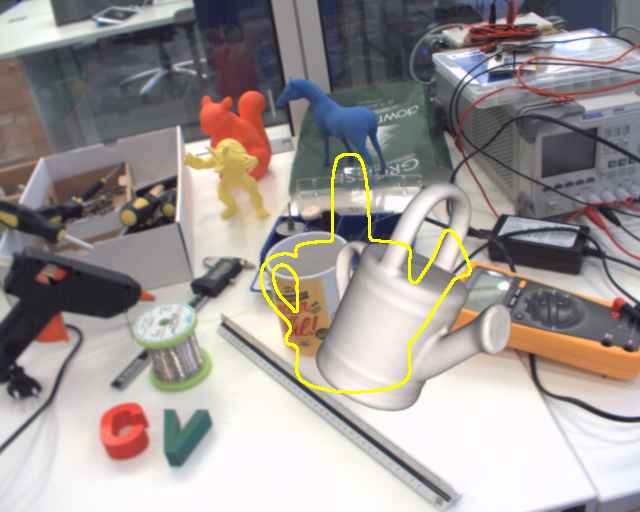} & \incfig{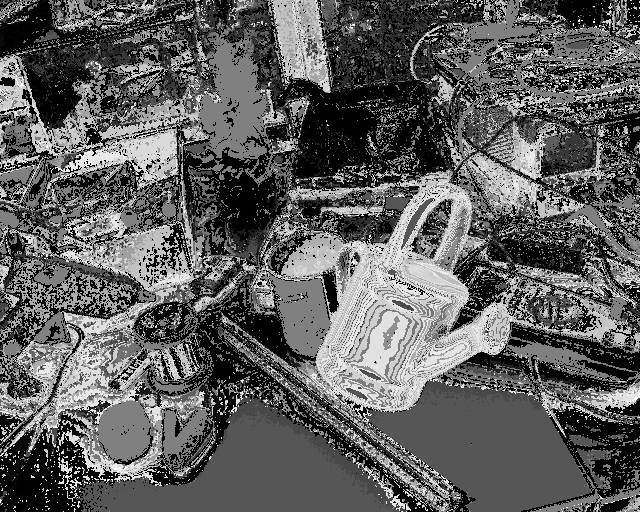} & \incfig{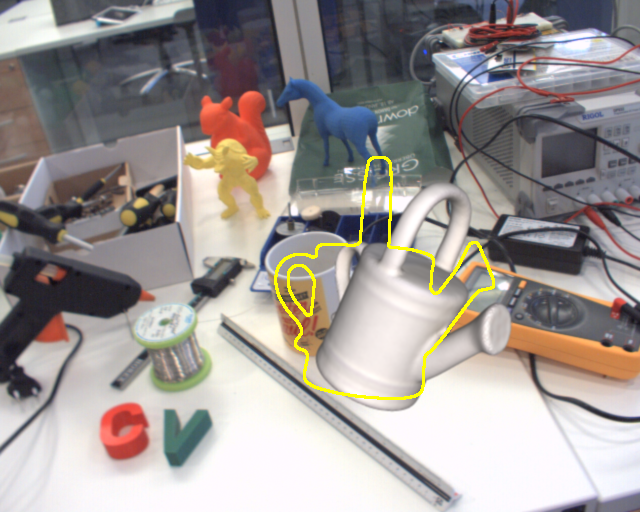} & \incfig{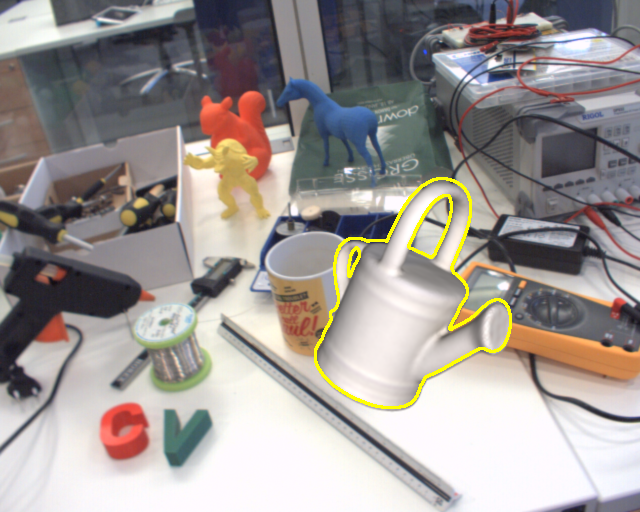} \\	
		(a) init.  & (b) srt3d~\cite{stoiber2021srt3d} & (c) ours prob. & (d) ours(local) & (e) ours
	\end{tabular}
	\caption{Visual examples and comparisons for $S=8$.}
	\label{fig:visresult}
\end{figure*}

\reffig{fig:visresult} shows some challenging examples with $S=8$. In the top row, the translation displacement is very large. Thanks to the use of long search lines, our local method can compensate for most translation errors and performs much better than SRT3D. In the bottom row, the rotation displacement is large and the probability map is very inaccurate, so both SRT3D and \ourslocal\, fail to converge properly. In both cases, our non-local method can result in the correct pose.


Due to the space limitation, results of other variants (\textit{dynamic light}, \textit{noisy}, \textit{occlusion}) are put in the supplementary material. The trend is generally the same as the \textit{regular} variant when compared with previous methods.


\subsection{Time Analysis}
\label{sec:timecost}

Runtime is measured on a machine with Intel(R) Core(TM) i7-7700K CPU. The pre-computation needs to be done for each of the $D$ directions ($D=16$ in our experiments). We implement it in parallel with the OpenCV \textit{parallel_for} procedure for acceleration,
which requires about $4\sim7$ms for each frame, depending on the object size. As a comparison, a sequential implementation requires $11\sim20$ms. 
After the pre-computation, the pose update iterations in \refeq{eq:delta} can be executed very fast. For $N=200$ as in our experiments, only about 0.03ms is required for each pose update iteration.

Besides the pre-computation, no parallelism is used in other parts of our code. For $S=1,2,3,4$, the average runtime per frame is about 10.8ms, 12.5ms, 21.7ms, 22.3ms respectively, achieving average about 50$\sim$100fps. The runtime varies a lot for different $S$ because of the pre-termination and near-to-far strategies in the non-local search. This is a desirable feature making our system adapt better to different displacements. On the contrary, previous non-local methods such as particle filter usually require constant computation regardless of the displacements.





\renewcommand\arraystretch{1.25}

\begin{table}[t]
	\setlength\tabcolsep{4pt}
	\begin{center}
		\caption{Ablation studies to the non-local search (GP=\textit{grid pretermination}, PP=\textit{path pretermination}, N2F=\textit{near-to-far search}). The results are computed with $S=4$ for the \textit{regular} variant. 	\textit{UpdateItrs} is the average pose update iterations per frame.}
		\label{tab:non-local-ablation}
		\begin{tabular}{ccccccc}
			\hline
			 & naive & +GP & +PP & +GP\&PP & +GP\&PP\&N2F & local(\ourslocal) \\ 
			\hline
			\textit{UpdateItrs} & 1085 & 718 & 823 & 578 & 468 & 59 \\
			\hline
			\textit{Time} & 47.7ms & 34.0ms & 37.9ms & 27.6ms & 22.3ms & 9.7ms \\
			\hline
			\textit{Accuracy} & 85.1 & 83.5 & 84.6 & 82.7 & 81.7 & 30.7 \\
			\hline
		\end{tabular}
	\end{center}
\end{table}

\subsection{Ablation Studies}
\label{sec:discussions}

Table \ref{tab:non-local-ablation} shows the ablation studies for the non-local search method. With the naive grid search, more than 1000 pose update iterations are required for each frame. Thanks to the proposed fast local optimization method, even the naive search can achieve near real-time speed. The three strategies all contribute significantly to higher efficiency. Using them all can reduce more than half of the time with only about 3\% sacrifice in accuracy. Moreover, compared with the local method, it increases the absolute accuracy by more than 50\% with only about $12ms$ sacrifice in time.

Table \ref{tab:alpha} shows the ablation experiments for the parameters $\alpha$ and $D$. 
%
%
As can be seen, $\alpha$ would take a great effect on the accuracy. Using $\alpha>1$ would significantly reduce the accuracy due to the erroneous contour correspondences. As shown in \reffig{fig:visresult}, the foreground probability map of some objects is very noisy.
Actually, for larger $\alpha$, the reduction of accuracy are mainly due to  the objects with indistinctive colors, as detailed in the supplementary material.

The number of directions $D$ would also take significant effects. Surprisingly, even with only the horizontal and vertical directions (i.e., $D=4$), our local method can still achieve 88.1\% accuracy, which is competitive in comparison with previous methods other than SRT3D.

Table \ref{tab:mn} shows the ablation experiments for the parameters $M$ and $N$. Our method is very insensitive to the choice of $M$. We set $M=3$ by default, but it is surprising that our method could benefit from using a larger $M$, which will increase the chance to find the correct correspondence, but at the same time introduce more noise. This further demonstrates the robustness of our method to erroneous contour correspondences.

The number of sampled contour points $N$ also takes some effects on the accuracy. From Table \ref{tab:alpha} we can find that $N$ is not difficult to set. For $N>200$, the increase of $N$ takes little effect on the accuracy. Note that the computation of our method is proportional to the number of sample points, so smaller $N$ may be considered for applications that are more time-critical.



\begin{table}[t]
	\setlength\tabcolsep{4pt}
	\begin{center}
		\caption{Ablation studies to $\alpha$, $D$ with \ourslocal\; and the \textit{regular} variant.}
		\label{tab:alpha}
		\begin{tabular}{ccccccc|cccccc}
			\hline
			$\alpha$ & 0.125 & 0.25 & 0.5 & 0.75 & 1.0 & 1.5 & $D$ &4& 8 & 12 &16 &20 \\ 
			\hline
			Acc. & 94.2 & 94.2 & 93.6 & 92.1 & 88.8 & 72.3 & Acc. & 88.1 & 92.8 & 93.7 & 94.2 & 94.3 \\
			\hline
		\end{tabular}
	\end{center}
\end{table}

\begin{table}[t]
	\setlength\tabcolsep{4pt}
	\begin{center}
		\caption{Ablation studies to $M$ and $N$ with the \textit{regular} variant.}
		\label{tab:mn}
		\begin{tabular}{ccccccc|cccccc}
			\hline			
			$M$ & 1 & 3 & 5 & 7 & 9 & 15 & $N$ & 50 & 100 & 200 & 300 & 400 \\
			\hline
			Acc. & 93.6 & 95.2 & 95.2 & 95.6 & 95.7 & 95.6 & Acc. & 93.9 & 94.6 & 95.2 & 95.3 & 95.2 \\
			\hline
		\end{tabular}
	\end{center}
\end{table}

\subsection{Limitations}

Our current method is still limited in some aspects that can be further improved. \textit{Firstly}, the error function $E'(.)$ is very simple, and it may be failed to recognize the true object pose in complicated cases. For example, for the case of \textit{noisy} variant and $S=1$, the mean accuracy is reduced from 83.4\% to 83.2\% after adding the non-local optimization (see the supplementary material). Obviously, this case should not occur if the error function is reliable.

\textit{Secondly}, the same as previous texture-less tracking methods, only the object shape is used for the pose estimation, so our method is not suitable for symmetrical objects (e.g., \textit{soda} in the RBOT dataset). For symmetrical objects, additional cues such as interior textures should be considered for improvements.

\textit{Thirdly}, since the correspondences are based on the probability map, the segmentation method would take a great effect on the accuracy of our method. Our current segmentation method is very simple considering the speed requirement, using better segmentation (e.g., those based on deep learning) would definitely improve the accuracy to a great extent.

\renewcommand{\incfig}[1]{\includegraphics[width=0.9\linewidth]{fig/6/#1.pdf}}

\section{Conclusions}

In this paper we proposed a non-local 3D tracking method to deal with large displacements. To our knowledge, this is the first non-local 3D tracking method that can run in real-time on CPU. We achieve the goal with contributions in both local and non-local pose optimization. An improved contour-based local tracking method with long precomputed search lines is proposed, based on which an efficient hybrid non-local search method is introduced to overcome local minimums, with non-local sampling only in the 2D out-of-plane space. Our method is simple yet effective, and for the case of large displacements, large margin of improvements can be achieved. Future work may consider extending the idea to related problems, such as 6D pose estimation~\cite{xiangPoseCNN2018} and camera tracking~\cite{zhang2021rosefusion}.

\subsubsection{Acknowledgements.}
This work is supported by NSFC project 62172260, and the Industrial Internet Innovation and Development Project in 2019 of China.

\clearpage
%
%
\bibliographystyle{splncs04}
\bibliography{egbib}
\end{document}